\documentclass[conference]{IEEEtran}
\IEEEoverridecommandlockouts
\usepackage[english]{babel}
\usepackage{blindtext}
\usepackage{tikz,pgfplots}
\usetikzlibrary{calc}
\usepackage{tkz-euclide}
\usepackage{bm}
\pgfplotsset{compat=1.5}
\usepgfplotslibrary{polar}
\usepackage{epstopdf} %converting EPS to PDF
\usepackage{amsmath}
\usepackage{wrapfig}
\usepackage{verbatim}
\usepackage{mathrsfs}
\usepackage{amssymb}
\usepackage{lipsum}  
\usepackage{acronym}
\usepackage{graphicx}
\usepackage{textcomp}
\usepackage{xcolor}
\usepackage{mathtools}
\usepackage{color,soul}
\usepackage{footnote}
\usepackage{booktabs}
\usepackage{multirow}
\usepackage{tikz}
\usepackage[font={small}]{caption}
\usepackage[font={small}]{subcaption}
\usepackage{algorithm}
\usepackage{algorithmic}
\usepackage{fancyhdr}
\usepackage{makecell}

\DeclareMathOperator*{\argmax}{arg\,max}

\usepackage{caption}
\usepackage{subcaption}
\usepackage{cleveref}
\usepackage{url}

\newcommand{\fig}[1]{Fig.~\ref{#1}}

\acrodef{ann}[ANNs]{Artificial Neural Networks}
\acrodef{adc}[ADC]{Analog-to-Digital Converter}
\acrodef{acc}[AC]{accumulate}
\acrodef{aoa}[AoA]{Angle of Arrival}
\acrodef{bp}[BP]{beam-pattern}
\acrodef{cae}[CAE]{Convolutional Autoencoder}
\acrodef{cir}[CIR]{Channel Impulse Response}
\acrodef{cnn}[CNN]{Convolutional Neural Network}
\acrodef{csnn}[CSNN]{Convolutional Spiking Layers}
\acrodef{ce}[CE]{Cross-Entropy}
\acrodef{dl}[DL]{Deep Learning}
\acrodef{dft}[DFT]{Discrete Fourier Transform}
\acrodef{fft}[FFT]{Fast Fourier Transform}
\acrodef{fmcw}[FMCW]{Frequency-Modulated Continuous-Wave}
\acrodef{har}[HAR]{Human Activity Recognition}
\acrodef{isac}[ISAC]{Integrated Sensing and Communication}
\acrodef{iqr}[IQR]{interquartile range}
\acrodef{lif}[LIF]{Leaky Integrate and Fire}
\acrodef{lse}[LSE]{Learned Spike Encoding}
\acrodef{lstm}[LSTM]{Long-Short Term Memory}
\acrodef {mmwave}[mmWave]{millimeter waves}
\acrodef{mse}[MSE]{Mean Squared Error}
\acrodef{mw}[MW]{Moving-Window}
\acrodef{mac}[MAC]{multiply–accumulate}
\acrodef{mud}[$\mu$D]{micro-Doppler}
\acrodef{ofdm}[OFDM]{Orthogonal Frequency Division Multiplexing}
\acrodef{rf}[RF]{Radio Frequency}
\acrodef{rd}[RD]{range-Doppler}
\acrodef{rmse}[RMSE]{Root Mean Squared Error}
\acrodef{sae}[SAE]{Spiking Autoencoder}
\acrodef{scae}[SCAE]{Spiking Convolutional Autoencoder}
\acrodef{sf}[SF]{Step-Forward}
\acrodef{snn}[SNN]{Spiking Neural Network}
\acrodef{snr}[SNR]{signal-to-noise ratio} 
\acrodef{ste}[STE]{Straight-Through Estimator} 
\acrodef{stft}[STFT]{Short-time Fourier Transform} 
\acrodef{tbr}[TBR]{Threshold-Based Representation}
\acrodef{uwb}[UWB]{Ultra-Wideband}

\begin{document}
% \bstctlcite{IEEEexample:BSTcontrol}

\title{Sparse Spike Encoding of Channel Responses for Energy Efficient Human Activity Recognition}

\author{Eleonora Cicciarella$^{*}$, Riccardo Mazzieri, Jacopo Pegoraro, Michele Rossi
\thanks{$^{*}$Corresponding author \texttt{eleonora.cicciarella@phd.unipd.it}. All authors are with the University of Padova, Department of Information Engineering.
This work was partially supported by the European Union under the Italian National Recovery and Resilience Plan (NRRP) Mission 4, Component 2, Investment 1.3, CUP C93C22005250001, partnership on “Telecommunications of the Future” (PE00000001 - program “RESTART”).}}

\IEEEoverridecommandlockouts
\IEEEaftertitletext{\vspace{-1\baselineskip}}
\maketitle
\begin{abstract}
ISAC enables pervasive monitoring, but modern sensing algorithms are often too complex for energy-constrained edge devices. This motivates the development of learning techniques that balance accuracy performance and energy efficiency. Spiking Neural Networks (SNNs) are a promising alternative, processing information as \textit{sparse} binary spike trains and potentially reducing energy consumption by orders of magnitude.

In this work, we propose a spiking convolutional autoencoder (SCAE) that learns tailored spike-encoded representations of channel impulse responses (CIR),  jointly trained with an SNN for human activity recognition (HAR), thereby eliminating the need for Doppler domain preprocessing. The results show that our SCAE-SNN achieves F1 scores comparable to a hybrid approach (almost 96\%), while producing substantially sparser spike encoding (81.1\% sparsity). We also show that encoding CIR data prior to classification improves both HAR accuracy and efficiency. The code is available at \texttt{https://github.com/ele-ciccia/SCAE-SNN-HAR}.
\end{abstract}


\section{Introduction}

\ac{har} has gained significant attention in recent years due to its wide range of applications in healthcare, smart homes, and related fields. Conventional \ac{har} solutions can be broadly classified into \emph{device-based} and \emph{device-free}. Device-based methods, such as wearables and smartphones, are effective but often impractical for long-term use. Device-free approaches, on the other hand, remove the burden of carrying sensors and instead rely on the environment. Vision-based systems are a common example, though their adoption is hindered by privacy concerns and sensitivity to lighting and temperature conditions.

These limitations have accelerated interest in \ac{rf}-based \ac{har}, which leverages the interaction between wireless signals and human movements~\cite{shastri2022review}. Since \ac{rf} signals are affected by reflection, diffraction, and scattering, they enable activity recognition without direct user involvement, thus preserving privacy. Moreover, \ac{rf}-based \ac{har} aligns with the emerging  \ac{isac} paradigm, where communication systems are extended with sensing capabilities. A key element in this setting is the \ac{cir}, which inherently captures variations in the wireless channel induced by human motions, offering a rich source of information for activity recognition.
In particular, the use of \ac{mmwave} \ac{isac} systems enhances sensing by providing higher temporal and spatial resolution through their large bandwidths and short wavelengths, resulting in clearer echo signatures and improved resolution of fine-grained human motions that are difficult to resolve at lower frequencies.

Due to the complexity of such channel, \ac{dl} solutions are commonly used to enable advanced sensing applications such as target recognition, movement analysis, and gesture recognition from Doppler spectrograms or range-Doppler maps~\cite{pegoraro2023rapid, ma2019wifi}. 
Although effective, these approaches incur high computational cost due to data volume and \ac{dl} inference, hindering the implementation on resource-constrained edge devices and routers, where \ac{rf} data are collected~\cite{lahmer2022energy}.                                                       

To overcome these challenges, \acp{snn}~\cite{eshraghian2023training} have recently gained attention for edge computing applications. Unlike standard \ac{ann} which process dense values through energy-hungry \ac{mac} operations, \acp{snn} encode information into \emph{sparse spike trains} and operate with lightweight \ac{acc} operations. Their event-driven processing offers orders-of-magnitude energy savings and has already been validated on neuromorphic hardware~\cite{frenkel20180}.

%As such, the way in which the input (channel) information is encoded into spikes is key in determining the trade-off between \textit{performance} on the learning task and \textit{sparsity} of the resulting spike train, which is tied to energy consumption.
In this work, we propose an end-to-end processing pipeline leveraging \acp{snn} to perform \ac{har} directly from raw CIR data, thus avoiding the need for heavy Doppler-domain preprocessing. Specifically, the main contributions are:
\begin{enumerate}
    \item We propose a \emph{learned} spike encoding tailored to \ac{cir} data, balancing activity recognition accuracy and spike sparsity. 
    Our \ac{snn} classifier trained on the proposed learned encoding achieves the highest F1 score ($\approx 96\%$) among competing approaches, including the \ac{cae}, \ac{snn} with no-encoding, and encoding via fixed delta thresholding. Moreover, our approach produces the sparsest spike representation, with $81.1\%$ sparsity compared to $28.6$\% and $71.7$\,\% for the \ac{cae} and delta thresholding methods, respectively. %highlighting the importance of learning the encoding for efficient spike-based processing.
    %\item We directly process raw \ac{cir} instead of Doppler images, thus substantially reducing signal processing overhead. 
    \item The proposed spike-based architecture effectively learns spectral features relevant to classification directly from \acp{cir}, whereas a standard deep \ac{cnn} originally trained on Doppler spectrograms experiences a performance drop of about 10 percentage points when trained on raw \ac{cir} data.
\end{enumerate}

The paper is organized as follows. Section~\ref{sec:rel_work} reviews related work on \ac{rf}-based \ac{har}. Section~\ref{subsec:background} introduces the necessary background on wireless channel modeling and the \ac{lif} spiking neuron model. The dataset and preprocessing steps are described in Section~\ref{sec:dataset}, while the proposed models and methods are detailed in Section~\ref{sec:methods}. Section~\ref{sec:results} presents and discusses the experimental results. Finally, Section~\ref{sec:conclusion} concludes the paper with a summary of findings and future research directions.

\section{Related work}
\label{sec:rel_work} 

Radar-based \ac{har} has relied on classical signal processing for feature extraction coupled with machine learning classifiers \cite{youngwook2009human}, or, more recently, on \ac{dl} architectures \cite{li2019survey}. To overcome the computational complexity and energy consumption of \ac{dl}, which hinders its deployment on resource-constrained devices, researchers have increasingly shifted to \acp{snn}. Their sparse and event-driven computations offer significant advantages in energy efficiency. 
One of the first studies in this direction is~\cite{banerjee2020application}, where the authors design a convolutional \ac{snn} to capture temporal dynamics in radar micro-Doppler signatures for \ac{har}. 
Subsequent research has largely focused on the more specific task of gesture recognition. 
%Wu et al. \cite{wu2025efficient} combine an enhanced Gaussian mixture model with an \ac{snn} with convolutional and recurrent units, successfully reducing computational complexity without compromising accuracy. 
The authors of~\cite{tsang2021radar} propose a recurrent \ac{snn} based on liquid state machines (LSMs) to classify \ac{rd} and \ac{mud} radar signals, using a signal-to-spike conversion scheme to encode them into spike trains. 
In \cite{arsalan2021radar}, a three-layer \ac{snn} with \ac{lif} neurons is utilized to recognize four hand gestures from \ac{rd} maps. 
The above works all share one main limitation, i.e., they rely on standard radar signal processing to obtain \ac{rd} or \ac{mud} maps, which are then encoded into spike sequences.
This approach does not fully exploit the sparsity of the radio channel features, since it processes high-dimensional tensors that mostly contain zeros at high computation and energy cost.
Conversely, in this paper, we make the \ac{snn} {\it learn directly from the raw \ac{cir}}, without inefficient preprocessing steps.

To the best of our knowledge, only works~\cite{arsalan2022spiking, cicciarella2024learned} have addressed \ac{har} from raw \ac{adc} or \ac{cir} data,
using a convolutional \ac{snn} and a hybrid \ac{cnn}-\ac{snn}, respectively.
However, in \cite{arsalan2022spiking} the \ac{snn} is partially constrained to mimic the \ac{dft} computation in standard radar processing, reducing its learning capabilities. Moreover, the \textit{spike activity} of the proposed network, i.e., the percentage of time in which the network neurons emit a spike, is not analyzed nor compared to other \ac{snn} models. On the contrary, \cite{cicciarella2024learned} requires a standard \ac{cnn} block to perform spike encoding, which shares the high energy consumption of standard \ac{dl}.
Conversely, we design an \textit{end-to-end} \ac{snn} comprising an encoder module that specifically learns to produce tailored spike trains to preserve human movement features, while minimizing spike activity for low energy operation.

\section{Background}\label{subsec:background}

\subsection{Wireless channel modeling}

The estimation of the \ac{cir} plays a central role in radar and \ac{isac} systems, as it embeds information about the location and movement of objects and people~\cite{pegoraro2023rapid}. The \ac{cir} is represented as a vector of complex channel gains, also referred to in the following as \emph{paths}, each characterized by an amplitude and a phase. Due to the finite delay resolution of the system, the \ac{cir} vector can only represent a discrete set of paths $\ell = 0, \dots, L-1$, each characterized by a propagation delay $\tau_\ell=\ell/B$, where $B$ denotes the transmission bandwidth and $1/B$ corresponds to the delay resolution. 
Assuming an analog beamforming system is used, multiple \ac{cir} estimations per packet can be obtained using different \acp{bp}.
The \ac{cir} obtained with each \ac{bp} encodes different environmental reflections, due to the different \ac{bp} shapes. 
Repeating the \ac{cir} estimation for each packet $k$ effectively samples the channel over time, with sampling period corresponding to the inter-packet transmission time $T_c$. For a given \ac{bp} $p$ and time instant (packet) $k$, the $\ell$-th component of the \ac{cir} can be written as
\begin{equation}\label{eq:cir}
    h_{\ell,p}[k] \triangleq h_{\ell,p}(kT_c) = a_{\ell,p}(kT_c)\,e^{j\phi_\ell (kT_c)}, 
\end{equation}
where $a_{\ell,p} (k)$ and $\phi_\ell(k)$ represent the complex gain and phase of path $\ell$ at time $k$, respectively. The path gain depends on the contribution of the \ac{bp} used for the transmission and on the reflectivity of the target, while the phase depends on the delay. 
%This formulation highlights why \ac{cir} estimation is valuable for \ac{isac}.  while conventional systems use it to configure transmission and decode data, in \ac{isac} the same multipath components reveal fine-grained environmental information. Wide mmWave bandwidths enable high delay resolution, allowing \ac{cir} to capture human-induced reflections. Temporal and angular variations in amplitude, delay, and phase provide motion, distance, and micro-Doppler cues, forming the basis for human sensing in \ac{isac} systems.

\subsection{Spiking neuron model}\label{subsec:background-snn_model}

Various spiking neuron models have been proposed that strike a good balance between biological plausibility and computational practicality. In this work, we adopt the \ac{lif} neuron model~\cite{gerstner2014neuronal}, which is widely used in learning applications due to its simplicity and fidelity in mimicking the behavior of biological neurons. Each \ac{lif} neuron is characterized by an internal state called \textit{membrane potential}, acting as a leaky integrator of the input signal. 
%Whenever the membrane potential exceeds a predefined threshold, the neuron emits an output spike and the membrane potential is instantaneously reset to a resting value. 
\ac{snn}s are constituted by a network of \ac{lif} neurons, typically organized into layers. Denoting by $U_i^{(\ell)}(t)$ the membrane potential of the $i$-th neuron in layer $\ell$ at time $t$, $R$ the input resistance, and $I_i^{(\ell)}(t)$ the input current to the neuron at time $t$, the dynamics of the membrane potential over time are described by the following differential equation:
\begin{equation}\label{eq_LIF}
    \tau \frac{dU_i^{(\ell)}(t)}{dt} = -U_i^{(\ell)}(t) + R I_i^{(\ell)}(t),
\end{equation}
where $\tau$ is the membrane time constant. Whenever the membrane potential exceeds a predefined threshold $\theta$, the neuron emits a unitary {\it spike} $S_i^{(\ell)}$, and $U_i^{(\ell)}$ is instantaneously reset to a resting value. 
Assuming $R=1\,\Omega$ and accounting for spiking and membrane potential reset, we formulate the above equation in discrete time using Euler's approximation method:
\begin{align} \label{eq_LIF_discrete}
    U_i^{(\ell)}[t] &= \beta\cdot U_i^{(\ell)}[t-1]  + I_i^{(\ell)}[t] - \theta\cdot S_i^{(\ell)}[t-1] \\
    I_i^{(\ell)}[t] &= \sum_j W_{ij}^{(\ell)}S_j^{(\ell-1)}[t]\\
    S_i^{(\ell)}[t] &= \Theta (U_i^{(\ell)}[t] -\theta), \label{eq_LIF_discrete-Heaviside}
\end{align} 
where $\beta = e^{-1/\tau}$ represents the decay rate of the membrane potential in the absence of input stimuli, and the input current is represented as a weighted sum of the incoming spikes. Eq.~(\ref{eq_LIF_discrete-Heaviside}) defines the spike generation mechanism, where $\Theta(\cdot)$ denotes the Heaviside step function. 
%The non-differentiability of the spikes makes \ac{lif} neurons unsuitable for gradient-based optimization. To address this issue, we adopt the \emph{surrogate gradient} method \cite{neftci2019surrogate}, where the gradient of the Heaviside step function is replaced with the gradient of a differentiable surrogate function during the backward pass. Throughout our work, we will use the surrogate fast sigmoid function \cite{zenke2018superspike}.

\section{Dataset} \label{sec:dataset}

\textit{Overview:} The DISC dataset~\cite{pegoraro2025disc} provides \ac{cir} measurements from standard-compliant IEEE 802.11ay packets, intended to support the evaluation of \ac{isac} methods. The \ac{cir} sequences capture signal reflections caused by human movement within a controlled, indoor environment characterized by complex multipath fading, due to the presence of furniture, computer equipment, display screens, and a large whiteboard. The data were collected using a $60$~GHz software-defined radio experimentation platform based on the IEEE~802.11ay Wi-Fi standard, which is not affected by frequency offsets due to its monostatic full-duplex operation mode.
The dataset comprises approximately $40$ minutes of \ac{cir} recordings, capturing reflections from 7 subjects performing $N_C=4$ different activities: \textit{walking}, \textit{running}, \textit{sitting down/standing up}, and \textit{waving hands}. 
%This part is characterized by uniform packet transmission times, with a granularity of over 3 \ac{cir} estimates per millisecond, yielding extremely high temporal resolution.
%Each CIR sample is a three-dimensional array with shape $(n_{rb}, n_p, n_{bp})$, where $n_{rb}$ represents the number of range bins (each bin corresponds to a specific time delay, i.e., a specific distance from the transmitter), $n_p$ is the number of packets transmitted in the sequence, and $n_{bp}$ corresponds to the number of \acp{bp} used per packet. 
See~\cite{pegoraro2025disc} for further details on the data acquisition process and experimental setup.

% \subsection{Dataset preprocessing}\label{subsec:dataset-preprocess}

\textit{Preprocessing:} Each record in the dataset represents a \ac{cir} sequence consisting of $110$ range bins, where each bin corresponds to a specific time delay, i.e., a specific distance from the transmitter, and the total length of the sequences may vary depending on the duration of the measurement.
Before being fed into the network, each complex-valued CIR sequence is segmented into partially overlapping time windows of length $W=64$ samples with step size $\delta=32$, providing short-term snapshots of the channel evolution. Within each window, the real and imaginary components are extracted, and the moving \ac{iqr} is computed along slow time to quantify temporal variability. The \acp{iqr} of the real and imaginary parts are combined into a joint measure, and the top $R=10$ range bins with the highest variability are selected, ensuring that only the most dynamic and informative portions of the \ac{cir} are retained as input features.
Both the real and imaginary parts of the \ac{cir} are preserved to fully retain the channel amplitude and phase information, the latter being particularly important as it encodes subtle variations induced by the human motion. Finally, each CIR is normalized to the range $[0,1]$ using min-max normalization.
%In particular, we apply a \ac{stft} using a Hamming window of $ W= 64$ samples and a stride of $\delta=32$ samples. Then, the absolute value is computed and only the top informative $R=10$ range bins are retained, discarding all the rest.

Notably, the information critical for classification is concentrated within about $2-3$ seconds of data, corresponding to approximately $N=232$ windows, when considering a sampling time $T_c=0.27$~ms. For this reason, sequences with a higher number of windows are divided into overlapping segments of length $N$, with an overlap of $N/2$ windows. This strategy enables the generation of additional samples from the same recorded sequence. Furthermore, it introduces temporal variability and partially incomplete activity patterns, which act as a regularizer during training and improve the robustness of the model to temporal uncertainty. Instead, sequences with a smaller number of windows are discarded, as they may not contain enough information to recognize the activity. 
As a result, after preprocessing, each sample has shape $(2, N, R, W)$, where the $2$ represents the real and imaginary components and will be treated as the input channel dimension for convolutional layers.

%To ensure a fair comparison, the division of the dataset into training, validation, and test was initially intended to follow the one adopted in \cite{pegoraro2025disc}, with subjects 1–5 assigned to the training set and subjects 6 and 7 to the validation and test sets, respectively. However, this configuration resulted in the absence of samples from the \emph{running} and \emph{waving hands} classes in the validation set, probably because the initial samples of the subjects were discarded because they were too short.
We included \ac{cir} sequences from subjects $1–4$ and subject $6$ in the training set, while subjects $5$ and $7$ are used for the validation and test sets, respectively. This ensures that every activity is represented at least once in each set and allows us to assess the generalization capability of the classifier to unseen subjects. The split yields $1,680$, $138$, and $182$ elements in the training, validation, and test sets, respectively.

\section{Models and methods} \label{sec:methods}

This section presents the proposed methodology. Our architecture consists of a \textbf{convolutional autoencoder} with \ac{lif} neurons, hereafter referred to as \emph{\ac{scae}}, used in combination with an \ac{snn}. The \ac{scae} is responsible for extracting a spike-based representation of the input data, while the \ac{snn} performs the recognition of four distinct human activities, acting as a classifier. These two components are {\it jointly trained} in an end-to-end fashion: the \ac{scae} is optimized to minimize the reconstruction error between the input \ac{cir} sample and its reconstruction, while the \ac{snn} is trained to maximize the classification accuracy. By adding the \ac{snn} block, we push the encoder to output a spike representation that preserves important information about the movement. A detailed description of the architecture is provided in Section~\ref{subsec:scae-snn}.

To evaluate the effectiveness of the proposed spike encoding, we compare it with  the following alternative strategies:
\begin{itemize}
    \item [a.] \emph{Convolutional autoencoder (CAE)}: this variant replaces the \ac{scae} block of the proposed architecture with a \ac{cae} with standard artificial neurons, as proposed in \cite{cicciarella2024learned} (Section~\ref{subsec:cae-snn}). 
    \item[b.] \emph{Delta thresholding encoding}: spike trains are generated using a thresholding mechanism (see Section~\ref{subsec:delta}) based on the temporal structure and variability of the input signal. Hence, such spikes are fed into our \ac{snn} classifier.
    \item[c.] \emph{Plain \ac{cir} processing (Direct-SNN)}: we also consider a baseline where no encoding is applied. \ac{cir} samples are directly fed into an \ac{snn} classifier, treating the input as a continuous current (see Section \ref{subsec:direct-snn}).
\end{itemize}

\subsection{SCAE-SNN architecture}
\label{subsec:scae-snn}

\begin{figure}[t!]
\centerline{\includegraphics[width=9.2cm]{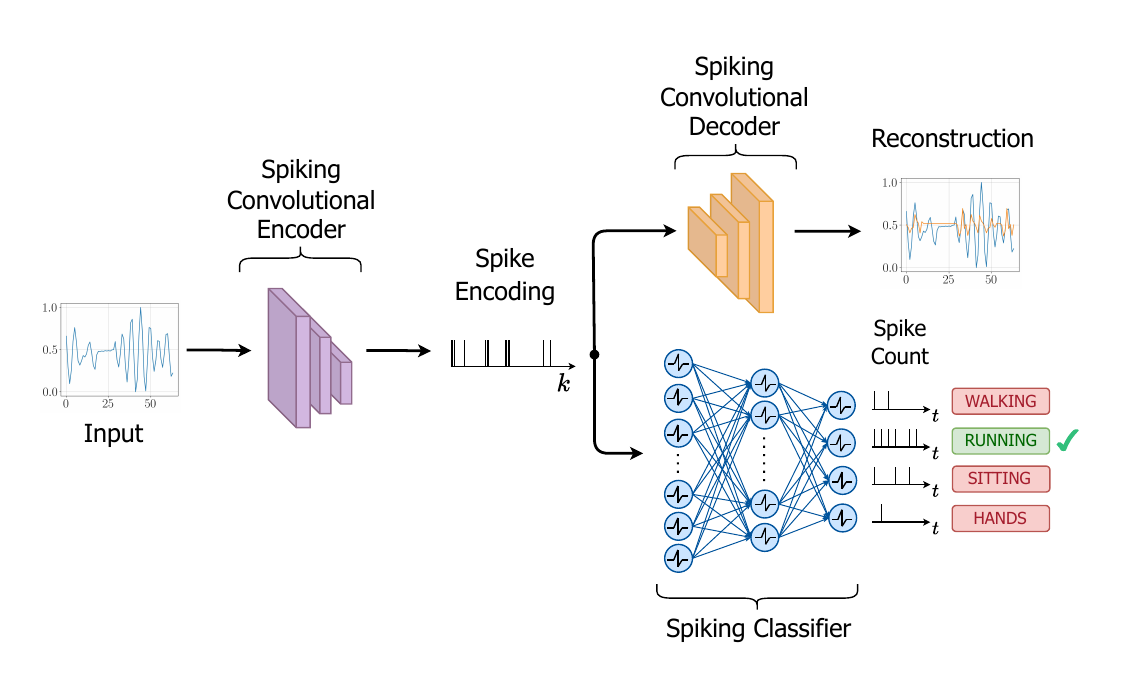}}
\caption{Block diagram of the proposed SCAE-SNN architecture.}
\label{fig:lse_architecture}
\vspace{-1\baselineskip}
\end{figure}

The proposed learning-based encoding method tailored to \ac{cir} samples is shown in \fig{fig:lse_architecture}. The structure of the encoder-decoder block (\ac{scae}) and the \ac{snn} follows the implementation described in \cite{cicciarella2024learned} with appropriate modifications.
In the following we provide a detailed description of each block.

\subsubsection*{SCAE}\label{subsubsec:scae}
The \ac{scae} is composed of an \textit{encoder} that takes as input the \ac{cir} samples and outputs the corresponding spike encoding. The latter is then fed into the \textit{decoder},  to reconstruct the original input.
In the proposed architecture, the encoder employs two three-dimensional convolutional layers (Conv3D), each of them followed by batch normalization and a dense layer of \ac{lif} neurons. The convolutional layers have $64$ and $2$ feature maps, respectively, and extract local spectral patterns in the input data using a kernel of dimension $1\times1\times3$. A stride of $1$ and zero padding are employed to preserve the input dimension. The \ac{lif} neurons are initialized with a decay factor $\beta=0.9$ and a threshold $\theta=1$. To provide greater flexibility, both parameters are then refined during training.
Although the encoder does not perform dimensionality reduction, it compresses the \ac{cir} by binarizing its values. This introduces information loss, preventing the autoencoder from simply learning the identity function.

The decoder block approximately inverts the operations performed by the encoder to reconstruct the \ac{cir} sample. It employs 2 transposed Conv3D layers with $64$ and $2$ feature maps, respectively. Similarly to the encoder, the first transposed Conv3D layer is followed by batch normalization and a layer of \ac{lif} neurons. In contrast, the second one uses a standard sigmoid activation, since the input is normalized within $[0,1]$. The remaining parameters are the same as those used in the encoder. Finally, the \ac{scae} is trained for $T=2$ timesteps, with each timestep processing a fixed sequence of $N/T = 116$ consecutive input windows.
%See Table~\ref{tab:cae_params} for a summary of the autoencoder parameters. 
% \begin{table}[h] 
% \centering
% \begin{tabular}{l c c}
% \hline
% \textbf{Parameter} & \textbf{Encoder} &\textbf{Decoder} \\ \hline
% Layer type & Conv3D & transposed Conv3D  \\ 
% N. of layers & 2  & 2 \\ 
% N. of filters  & 64 - 2 & 64 - 2 \\ 
% Kernel size & (1,1,3) &  (1,1,3) \\
% Stride & 1 & 1 \\
% Zero padding & yes & yes \\
% Decay rate $\beta$  & 0.9 & 0.9 \\
% Threshold $\theta$  & 0.8 & 0.8 \\
% \hline
% \end{tabular}
% \caption{Parameters of the SCAE architecture. All convolutional layers share the same kernel, stride, and padding settings. Coefficients $\beta$ and $\theta$ of \ac{lif} neurons are initialized with the values reported in the table.}
% \label{tab:cae_params}
% \end{table}

\subsubsection*{SNN}\label{subsubsec:snn}
The \ac{snn} classifier receives the spike encoding and predicts human activity. 
The encoding is first downsampled by average pooling (kernel $1\times1 \times4$ and stride $1\times2\times2$), then processed by three fully connected layers with $128$, $64$, and $N_C$ \ac{lif} neurons, respectively. The decay factor $\beta$ and the firing threshold $\theta$ are initialized with values $0.9$ and $0.8$, respectively, and are refined during training, except for the final classification layer which uses a fixed $\theta=1$. The network is trained for $29$ timesteps and the output spikes are interpreted using a \emph{rate coding} scheme (further detailed in Section \ref{subsec:train}). %Specifically, since the total number of windows is $N=232$, at each timestep the network processes $N/29 = 8$ consecutive windows.

\subsection{CAE-SNN architecture}\label{subsec:cae-snn}

The \ac{cae} produces \emph{bipolar} spikes (both positive and negative) using a custom Heaviside function centered at $\tau=0.4$, as described in \cite{cicciarella2024learned}. 
%\blue{The possible benefit of this hybrid approach is that bipolar spikes can potentially improve the representational capacity of the encoded information. Moreover, the \ac{cae} processes the input in a single timestep, in an attempt of reducing the latency.}
%However, we argue that, given the temporal structure of the input data, an autoencoder with \ac{lif} neurons is better suited to capture temporal dependencies. 
%Importantly, although a spiking network typically processes the input over multiple timesteps, the associated overhead is offset by their highly sparse computation.
The \ac{cae} shares the same structure and parameters as the \ac{scae}, except for the \ac{lif} neurons that are replaced by ReLU activation, and is jointly trained with the \ac{snn} previously described to minimize the reconstruction error and maximize the classification accuracy.

\subsection{Delta thresholding encoding}\label{subsec:delta}

Letting $n$ be the n-th time sample, and $m$ the m-th range bin, the delta filter for a single \ac{cir} sample is 
\begin{equation}
    x_\delta (n,m) = x(n,m)-x(n-1,m),
\end{equation}
where this equation is independently applied to the real and imaginary parts of the \ac{cir}. 
Calling $\sigma_i$ the variance of $x_\delta(i,m)$, for each $i=1,\dots,W$, and $\alpha=\mathbb{E}[\sigma_i]$, the final spike encoding $s$ is obtained by thresholding the $x_\delta$ as follows:
\begin{equation}
    s(n,m) =
\begin{cases}
    1, &\text{if } x_\delta(n,m)>\alpha\\
    -1, &\text{if } x_\delta(n,m)<-\alpha\\
    0,  &\text{otherwise.}\\
\end{cases}
\end{equation}
Despite its simplicity, this method has been proven to be effective for \ac{rf} data \cite{mueller2023aircraft}. For a fair comparison of the different encoding strategies, the \ac{snn} of Section~\ref{subsec:scae-snn} is trained on the resulting spike trains to recognize human activities.

\subsection{Direct-SNN}\label{subsec:direct-snn}
Two types of \acp{snn} are evaluated for direct classification from \acp{cir}. The first uses fully connected layers with either three ($64$–$128$–$N_c$) or four ($64$–$128$–$256$–$N_c$) layers, where input features are downsampled via average pooling as in Section~\ref{subsec:scae-snn}. The second type employs convolutional feature extractors with one, two, or three layers (kernel $1\times1\times3$, stride $1$), followed by a linear readout, using the channel progression $64 \rightarrow 128 \rightarrow  256$. All models share the same number of timesteps ($T=29$) and parameter configuration for $\beta$ and $\theta$ as the \ac{snn} described in Section~\ref{subsec:scae-snn}. Section \ref{sec:results} reports the best results for each type: \emph{Direct-SNN-Lin} with four linear layers and \emph{Direct-SNN-Conv} with three convolutional layers.

\subsection{Training method}\label{subsec:train}

Next, we describe the training process for each architecture, starting with the output decoding strategy.

\subsubsection*{Output spike decoding}
In this work, we interpret output spikes as rate encoded outputs. Specifically, let $N_C$ denote the number of output classes, and let $\boldsymbol{S}[t]\in\{0,1\}^{N_C}$ be the spike train emitted by the output layer of the \ac{snn} at timestep $t$. We compute the spike count vector $\boldsymbol{c}\in\mathbb{Z}^{N_C}$ by summing the spike trains over $T$ time steps: $\boldsymbol{c}=\sum_{t=0}^{T-1} \boldsymbol{S}[t]$. The predicted class label $\hat{\boldsymbol{y}}$ is then determined by the index of the neuron with the highest spike count: $\hat{\boldsymbol{y}} = \argmax_i c_i$.
Rate decoding offers error tolerance by distributing the information across multiple spikes, reducing the impact of individual spike failures on the final prediction. Additionally, the presence of more spikes enhances learning dynamics by providing a stronger gradient signal during backpropagation~\cite{eshraghian2023training}.
%thereby mitigating issues such as the dead neuron problem \cite{eshraghian2023training}.

\subsubsection*{Loss function}
Both \ac{scae}-\ac{snn} and \ac{cae}-\ac{snn} architectures are trained end-to-end through the minimization of a loss function that depends on the output of the decoder and of the \ac{snn}. 
Specifically, let $\boldsymbol{X}$ and $\hat{\boldsymbol{X}}$ be the input \ac{cir} measurement and its reconstruction produced by the decoder, respectively. We define the reconstruction loss of the autoencoder using the \ac{mse} as $\mathcal{L}_{\rm rec} \triangleq \mathrm{MSE}(\boldsymbol{X}, \hat{\boldsymbol{X}})$. Moreover, let $\boldsymbol{y}$ be the correct class and $\boldsymbol{c}$ be the output spike count, as defined above. Then, we define the classification loss using the \ac{ce} Spike Count as 
\begin{equation}
    \mathcal{L}_{\rm class} = \sum_{i=0}^{N_C} y_i \log(p_i),\quad \mathrm{with}\;\;  p_i=\frac{e^{c_i}}{\sum_{j=1}^{N_C} e^{c_j}}.
\end{equation}
Hence, the final loss for training both \ac{scae}-\ac{snn} and \ac{cae}-\ac{snn} is given by $\mathcal{L} = \mathcal{L}_{\rm rec} + \gamma\, \mathcal{L}_{\rm class}$, with $\gamma$ being a scaling parameter.
 All Direct-\acp{snn} architectures are trained by minimizing $\mathcal{L}_{\rm class}$ only. 
 
 To address class imbalance, we introduce weights in $\mathcal{L}_{\rm class}$, chosen in $(0,1]$ and selected via grid search. The weights are set inversely proportional to the class frequency and the classification difficulty, assigning lower values to common or easily recognized classes. Specifically, \emph{walking} and \emph{running} share a weight of $0.3$, \emph{sitting down} is assigned $0.6$, and the most challenging class, \emph{waving hands}, receives $1$. Indeed, the latter only involves arm movements, which do not cause strong scattering of the radio signal and are hence difficult to detect.

\subsubsection*{Training and hyperparameter optimization}

All architectures are trained with batches of $8$ \ac{cir} samples, using backpropagation with Adam optimizer (learning rate $\eta=10^{-4}$). To address the non-differentiability of the spike generation, we adopt the \emph{surrogate gradient} approach~\cite{neftci2019surrogate} with the arctangent approximation of the firing threshold during the backward pass. Training is performed for a maximum of $50$ epochs and is stopped if the validation F1 score does not improve for $15$ consecutive epochs.
Network hyperparameters are optimized through a grid search over a limited set of values, constrained by available resources, using validation loss as the objective. For the $\tau$ parameter of \ac{cae}, we performed five runs per candidate value to compute the mean and standard deviation of the encoding sparsity and classification accuracy on the validation set. Finally, we set $\gamma=1$ in all experiments, as empirical observations showed that $\mathcal{L}_{\rm class}$ consistently remained small compared to $\mathcal{L}_{\rm rec}$, likely due to the high sparsity of spike-encoded data.

\section{Experimental results}\label{sec:results}
In this section, we compare the spike encodings produced by \ac{scae}, \ac{cae}, and delta thresholding for human activity recognition across multiple metrics, including classification accuracy and encoding sparsity. Additionally, we benchmark these methods against \emph{Direct-\ac{snn}}, where CIR samples are fed directly into the \ac{snn} classifier, and against the six-layer \ac{cnn} from~\cite{pegoraro2025disc} originally trained on Doppler spectrograms and retrained here on raw \ac{cir} data for a fair comparison. Computational efficiency is evaluated in terms of model size and average inference latency over ten runs, while spike activity rate serves as a proxy for energy consumption in neuromorphic hardware. These metrics provide insights into both effectiveness and practicality for deployment in resource-constrained environments. All the code is developed in Python, using the PyTorch and the snnTorch~\cite{eshraghian2023training} frameworks.

%In this section, we compare the effectiveness of the spike encoding produced by \ac{scae}, \ac{cae}, and delta thresholding for recognition of human activities across multiple performance metrics.
%Specifically, we assess each encoding strategy in terms of classification accuracy and encoding sparsity, providing insights into effectiveness and efficiency. We also benchmark these methods against ``direct \ac{snn}'', where \ac{cir} samples are direclty fed into the \ac{snn} classifier. 
%In addition, we analyze the computational efficiency of the resulting architectures in terms of model size and inference time, measured as the average latency required for a single inference task, averaged over ten runs. 
%For \ac{snn} architectures, we also obtain the average spike activity rate given the encoding, which serves as a proxy for energy consumption in neuromorphic hardware devices. 
%These metrics provide a broader perspective on the practical feasibility of deploying each architecture, particularly in resource-constrained environments. All the code is developed in Python, using the PyTorch and the snnTorch frameworks \cite{eshraghian2023training}.

\subsection{Accuracy results}\label{subsec:acc-results} 

\begin{figure}[t!]
\centerline{\includegraphics[width=7.5cm]{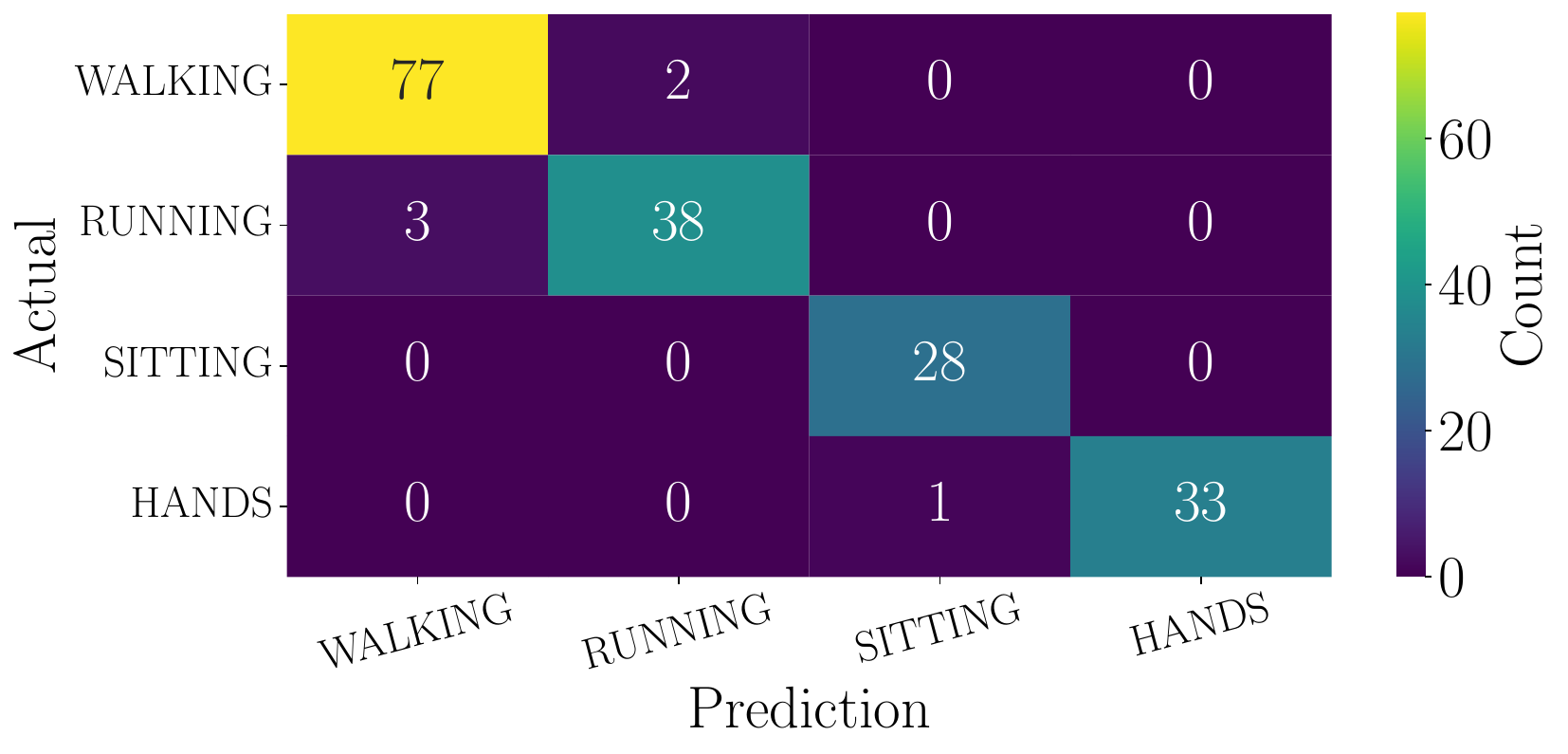}}
\caption{Confusion matrix of the SCAE-SNN model on the test set.}
\label{fig:confusion_mx}
\vspace{-1\baselineskip}
\end{figure}

We compare the performance of the models on the test set using the \emph{F1 score}, as it offers a more stable and reliable evaluation of the model performance when dealing with multiclass classification tasks over imbalanced datasets. 
To aggregate performance across classes, we employ the \emph{macro-averaging} strategy, which calculates the F1 score independently for each class and then computes the arithmetic mean, ensuring equal contribution from all classes and a fair assessment of underrepresented ones.
As shown in Table \ref{tab:results}, the \ac{scae}-\ac{snn} and \ac{cae}-\ac{snn} models achieve comparable performance with F1 scores exceeding $95\%$, demonstrating their ability to generalize to unseen subjects (see Figure~\ref{fig:confusion_mx}). The \ac{snn} trained on delta-threshold encoded data reaches approximately $75\%$ F1 score, suggesting that a simple fixed threshold encoding is not sufficient for effective radar-based \ac{har}. In contrast, direct training of an \ac{snn} on raw \ac{cir} samples results in markedly degraded performance (e.g., the Direct-SNN-Conv consistently predicts a single class) indicating limited capacity to extract discriminative spectral features. Furthermore, the baseline \ac{cnn} from~\cite{pegoraro2025disc} exhibits a reduction of roughly ten percentage points in F1 score when retrained on raw \ac{cir} data compared to the results reported in the original paper, highlighting the superiority of the proposed spike-based architecture in learning directly from raw signals.
 
% \begin{table}[ht!]  
%   \centering
%   %\captionsetup{width=.73\linewidth}
%   \scalebox{0.95}{
%   \begin{tabular}{l|c c c c}
%     \hline
%     \textbf{Method} & \textbf{F1 score~[\%]} & \textbf{Params} &
%     \makecell{\textbf{Inference}\\\textbf{time~[ms]}} &
%     \makecell{\textbf{Spike}\\\textbf{rate~[\%]}} \\
%     \hline
%     SCAE-SNN &  93.68 $\pm$ 0.03 & 28,442 &  1.2 + 18.4 & 5.25\\
%     CAE-SNN  &  95.19 $\pm$ 0.01 & 28,437 &  0.6 + 18.6 & 6\\
%     Delta thr. + SNN &  75.61 $\pm$ 0.04 & 26,505 & 18.7 & 9.3 \\
%     Direct-SNN-Lin & 0.514 & 67,723 &  24.6 &4\\
%     Direct-SNN-Conv & 0.414 &  2,646,985 &  21.9 & 35.1\\
%     Baseline \cite{pegoraro2025disc} & 0.875 & 770,028 & 0.9 & / \\
%     \hline
%   \end{tabular}
%   }
%   \caption{Comparison of classification and efficiency metrics on test set. We also include the results of the baseline from \cite{pegoraro2025disc}, which is a six-layer standard \ac{cnn} retrained on our CIR samples.}
%   \label{tab:results}
% \end{table}

\begin{table}[t!]  
  \centering
  %\captionsetup{width=.73\linewidth}
  \scalebox{0.92}{
  \begin{tabular}{l|c c c c}
    \hline
    \textbf{Method} & \textbf{F1 score~[\%]} & \textbf{Params} &
    \makecell{\textbf{Inference}\\\textbf{time~[ms]}} &
    \makecell{\textbf{Spike}\\\textbf{rate~[\%]}} \\
    \hline
    SCAE-SNN &  $\mathbf{95.75 \pm 1.18}$ & $2.8 \times 10^4$ & $ 1.2 + 18.6$ & $10.7$\\ %OK
    CAE-SNN  &  $95.16 \pm 2.35$ & $2.8 \times 10^4$ &  $0.7 + 18.7$ & $11.9$\\
    Delta thr. + SNN &  $74.59 \pm 3.67$ & $2.6 \times 10^4$ & $18.8$ & $8.0$ \\%OK
    Direct-SNN-Lin & $49.98 \pm 3.83$ & $6.7 \times 10^4$ &  $24.4$ &$8.5$\\ %OK
    Direct-SNN-Conv & $19.45 \pm 8.64$ &  $2.5 \times 10^6$ &  $27.9$ & $26.2$\\
    Baseline CNN \cite{pegoraro2025disc} & $85.87  \pm 4.47$ & $7.7 \times 10^5$ & $0.5$ & / \\
    \hline
  \end{tabular}
  }
  \caption{Performance comparison on the test set. F1 scores are reported as mean $\pm$ standard deviation over five weight initializations.}
  \label{tab:results}
\end{table}

%\textbf{Classe HANDS}: discutere della complessità della classe, spiegare brevemente quale potrebbe essere il motivo e dire che il nostro metodo raggiunge recall massimo.

\subsection{Sparsity of the encoding}\label{subsec:sparsity}

We define the sparsity of a spike encoding as the fraction of zeros it contains. The sparsity directly correlates with the efficiency of the \ac{snn}: higher sparsity implies fewer computations, which translates into a lower energy consumption. 
%As shown in Table~\ref{tab:sparsity}, our \ac{scae} achieves the highest sparsity (over $81$\%) compared to standard \ac{cae} (almost $29\%$), and to delta thresholding (almost $72$\%). These results suggest that our spike encoding can significantly reduce computational load and energy usage while maintaining effective representations.
As shown in Table~\ref{tab:sparsity}, our \ac{scae} produces the sparsest spike encoding (over $81$\%), being approximately $2.8\times$ sparser than the standard \ac{cae} ($\approx29$\%) and notably sparser than the delta thresholding approach ($\approx72$\%). This result indicates that our spike encoding can significantly reduce computational load and energy usage, while maintaining effective representations.

\subsection{Computational efficiency}\label{subsec:energy}

\begin{table}[t!]
\centering
\begin{tabular}{l c}
\hline
\textbf{Encoding method} & \textbf{Sparsity} \\ \hline
SCAE & $81.1$\% \\ 
CAE  & $28.6$\%  \\ 
Delta Thresholding  &  $71.7$\%\\ \hline
\end{tabular}
\caption{Average sparsity of the encoding methods on the test set.}
\label{tab:sparsity}
\end{table}

Our \ac{scae}-\ac{snn} uses a compact architecture with 28\,k parameters, making it one of the smallest models considered. 
%The reported inference time for the \ac{scae}-\ac{snn} architecture is given as the sum of the encoding and classification times, totaling approximately $19.8$\,ms, which is slightly higher than the \ac{cae}-\ac{snn} but still substantially lower than the Direct-SNN approach. This shows that the inclusion of the encoding stage does not significantly affect overall latency. 
For the first two architectures, the reported inference time in Table~\ref{tab:results} corresponds to the sum of the encoding and classification stages, whereas other methods report only the classification time. The \ac{scae}-\ac{snn} achieves a total inference time of approximately $19.8$\,ms, which is slightly higher than the \ac{cae}-\ac{snn}, yet substantially lower than the Direct-SNN approaches, which perform only classification.
Nevertheless, the higher encoding time is compensated for by the \ac{scae} event-driven operation and high sparsity, which leads to fewer effective computations on neuromorphic hardware. In terms of spike activity, our method exhibits a firing rate of $10.7\%$, lower than \ac{cae}-\ac{snn}, confirming its computational efficiency and potential for energy savings.
%In terms of spike activity, our method exhibits a low firing rate of $5.25\%$, indicating efficient event-driven computation and energy savings.
%Conversely, the Direct-SNN-Conv network shows a much higher activity rate of $35.1\%$, mainly driven by the average spike rate in the last layer, highlighting that the network struggles to correctly classify the data. 

%we wanted to have have compact architecture which facilitates potential deployment on resource-constrained devices).

\subsection{Tradeoff between F1 score and inference time}
%To better understand the impact of key components on the performance of our architecture, we conduct a series of ablation studies. Specifically, we analyze how variations in the firing threshold, decay factor, and number of timesteps affect the performance and temporal dynamics. 

%\subsection{Firing threshold}
%The firing threshold is a crucial hyperparameter, since it is tied to the spike activity rate. A lower threshold leads to more frequent spiking activity, potentially improving sensitivity to variations in input stimuli but also increasing noise and energy consumption. Conversely, a higher threshold sparsifies the activity, which may improve efficiency but may result in under-activation. Therefore, adjusting the firing threshold directly influences the trade-off between responsiveness and sparsity.

%\subsection{Decay rate}
%The decay factor controls the rate at which the membrane potential decays over time in the absence of input stimuli. A strong decay accelerates the loss of past information, making the neuron more responsive to recent inputs but less able to integrate over longer temporal windows. A weaker decay allows information to persist for longer, facilitating temporal integration but potentially leading to overlapping activations.

\begin{figure}[t!]
\centerline{\includegraphics[width=7.5cm]{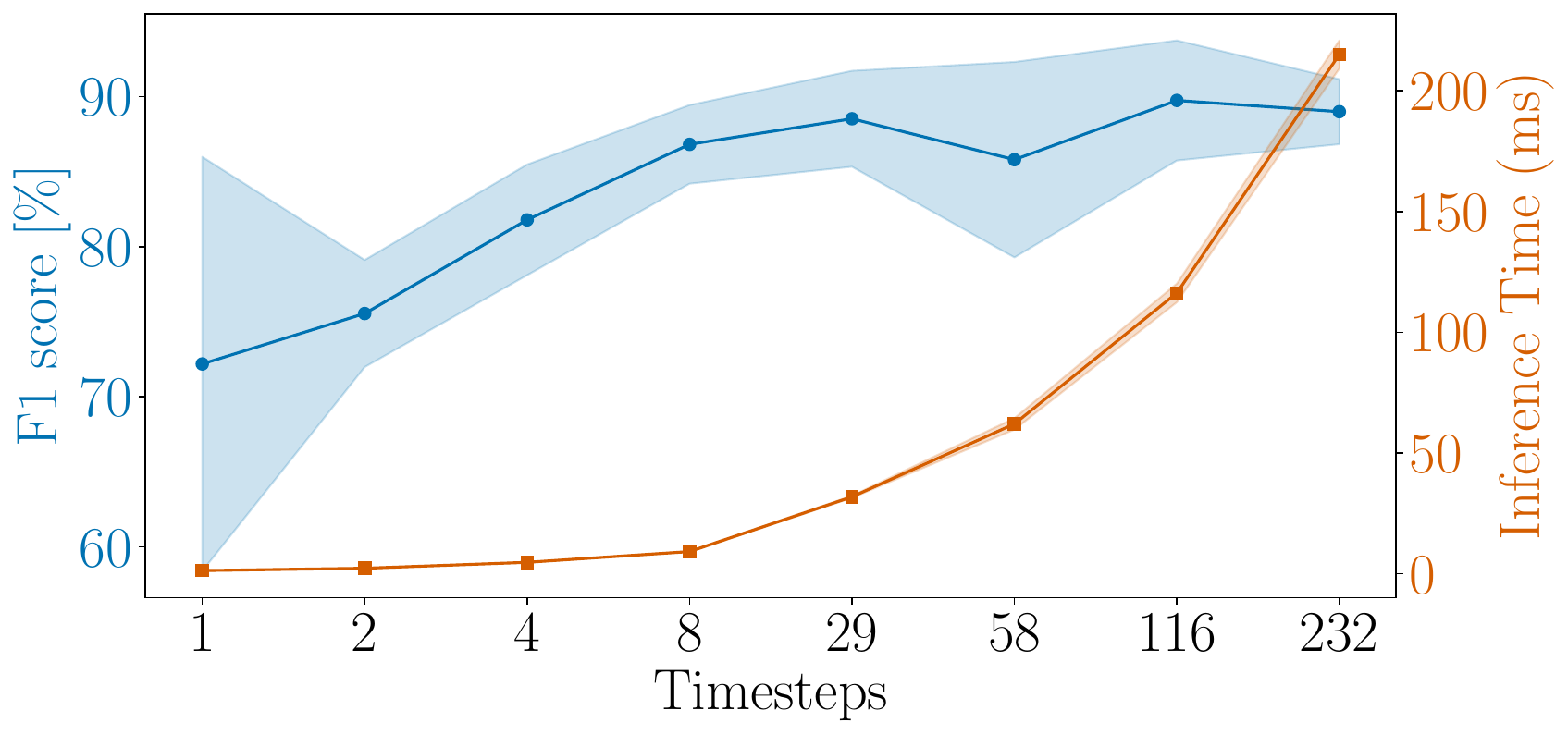}}
\caption{Inference time vs. F1 score across timesteps. Shaded areas represent $\pm$ one standard deviation computed over ten seeds.}
\label{fig:ablation_timestep}
\vspace{-1\baselineskip}
\end{figure}

%The inference time $T$ (expressed in number of timesteps) determines the temporal resolution available for processing input sequences. 
The classification inference time is directly linked to the number of timesteps $T$ and defines the temporal resolution of the model. Increasing $T$ enables richer temporal dynamics and improved accuracy in capturing sequential dependencies, but at the cost of higher computational overhead. Reducing it improves efficiency, but may limit the network ability to fully exploit temporal information. 
Figure~\ref{fig:ablation_timestep} shows the tradeoff between the classification inference time and the F1 score. As expected, a larger $T$ leads to an increase in the F1 score, which however, tends to saturate after $29$ timesteps. At the same time, the inference time grows proportionally, as the \ac{snn} must process the input over more timesteps. These results suggest that $T=29$ represents a suitable choice for runtime deployment of the architecture.

\section{Concluding remarks}\label{sec:conclusion}

In this paper, we proposed an end-to-end pipeline for radar-based \ac{har} using \acp{snn}, introducing a learned spike encoding tailored to raw \ac{cir} data, thus avoiding the need for costly Doppler domain preprocessing. %The proposed architecture consists of a spiking convolutional autoencoder that learns to generate sparse spike trains preserving human movement features while minimizing spike activity for energy efficiency, and an \ac{snn} classifier that performs activity recognition directly from the encoded \ac{cir} data. 
The proposed architecture consists of a spiking convolutional autoencoder combined with an \ac{snn} classifier. The first learns to generate sparse spike trains that preserve human movement features, while minimizing spike activity for energy efficiency. The latter, instead, performs activity recognition directly from the encoded \ac{cir} data. The modules are jointly trained to balance reconstruction fidelity and classification accuracy. 
Experimental results demonstrate that the proposed approach achieves high efficiency in terms of model size, inference time, and spike activity, without compromising accuracy, making it a practical solution for real-time deployment on resource-constrained devices. Future research will focus on extending the framework to multi-antenna systems, enabling richer spatial sensing, as well as evaluating the approach in multi-user \ac{isac} scenarios.
\vspace{-0.4mm}

\bibliography{references}
\bibliographystyle{ieeetr}
\end{document}